\documentclass[runningheads]{llncs}
\usepackage[T1]{fontenc}
\usepackage{graphicx}
\usepackage{float}
\usepackage{subcaption}
\usepackage{multirow}
\usepackage{xcolor}
\usepackage{hyperref}
\usepackage{comment}
\usepackage{siunitx}
\usepackage{tablefootnote}
\usepackage{gensymb}

\begin{document}
\title{A Discussion About Violin Reduction: Geometric Analysis of Contour Lines and Channel of Minima}
\author{Philémon Beghin\inst{1,2}\orcidID{0000-0003-0354-7901} \and \\
Anne-Emmanuelle Ceulemans \inst{1,3,4}\orcidID{0000-0002-3273-4004} \and \\
François Glineur \inst{1,2}\orcidID{0000-0002-5890-1093}}
\authorrunning{Ph. Beghin et al.}
\institute{UCLouvain, Louvain-la-Neuve, Belgium \and
Institute of Information
and Communication Technologies, Electronics and Applied Mathematics (ICTEAM), Louvain‐la‐Neuve, Belgium \and
Institute for the Study of Civilisations, Arts and Letters (INCAL), Louvain-la-Neuve, Belgium \and Musical Instruments Museum (MIM), Brussels, Belgium
\email{\{philemon.beghin,anne-emmanuelle.ceulemans,francois.glineur\}@uclouvain.be}}
\maketitle              % typeset the header of the contribution
\begin{abstract}
Some early violins have been reduced during their history to fit imposed morphological standards, while more recent ones have been built directly to these standards. Geometric differences exist between reduced and unreduced instruments, particularly in their contour lines and channel of minima. In a recent preliminary work, we computed and highlighted those two features for two instruments using triangular 3D meshes acquired by photogrammetry, whose fidelity has been assessed and validated with sub-millimetre accuracy. We propose here an extension to a corpus of 38 violins, violas and cellos, and introduce improved procedures, leading to a stronger discussion of the geometric analysis. We first recall the material we are working with. We then discuss how to derive the best reference plane for the violin alignment, which is crucial for the computation of contour lines and channel of minima. Finally, we show how to compute efficiently both characteristics and we illustrate our results with a few examples. \\

\end{abstract}

\vspace{0.3 cm}

\section{Motivation and Related Works}
\label{sec:Introduction}

\noindent The origins of the violin, in the broadest sense, date back to the end of the sixteenth century. The instrument immediately gained in popularity and spread throughout Europe and around the world. This wave of success gave rise to a very broad family of instruments. Luthiers experimented with numerous techniques, depending on the influence of their masters or families, their era and their geographical area, before converging on reference shapes and sizes adopted around 1750 \cite{ceulemans2023baroque}. These sizes, still used nowadays\footnote{The violin, viola, cello and double-bass are the four instruments of the violin family. We will focus here only on the first three. However, four instruments from our corpus stand out slightly from the rest, described in Appendix \ref{sec:Appendix}, and are called tenor violins and bass violins. These are names and sizes used before the standardisation of instruments, around 1750. The tenor violins are slightly larger than violas, and the bass violins are slightly larger than cellos.}, stem from a desire to standardise instruments, which marked the transition from the baroque to the classical period. Instruments that did not correspond to a standard size were reduced (the reverse operation was not possible, or at least extremely rare \cite{echard2013documentary}). Today, it is important for musicologists to determine whether or not an instrument has been reduced and, if so, to quantify how. In our research, we aim to objectively differentiate between reduced and unreduced instruments, which is currently only done through the empirical expertise of luthiers and specialists. Our objective approach therefore complements this observation work.  \\

\begin{figure}[H]
  \centering
  \begin{subfigure}{0.27\linewidth}
    \centering
    \includegraphics[scale = 0.13]{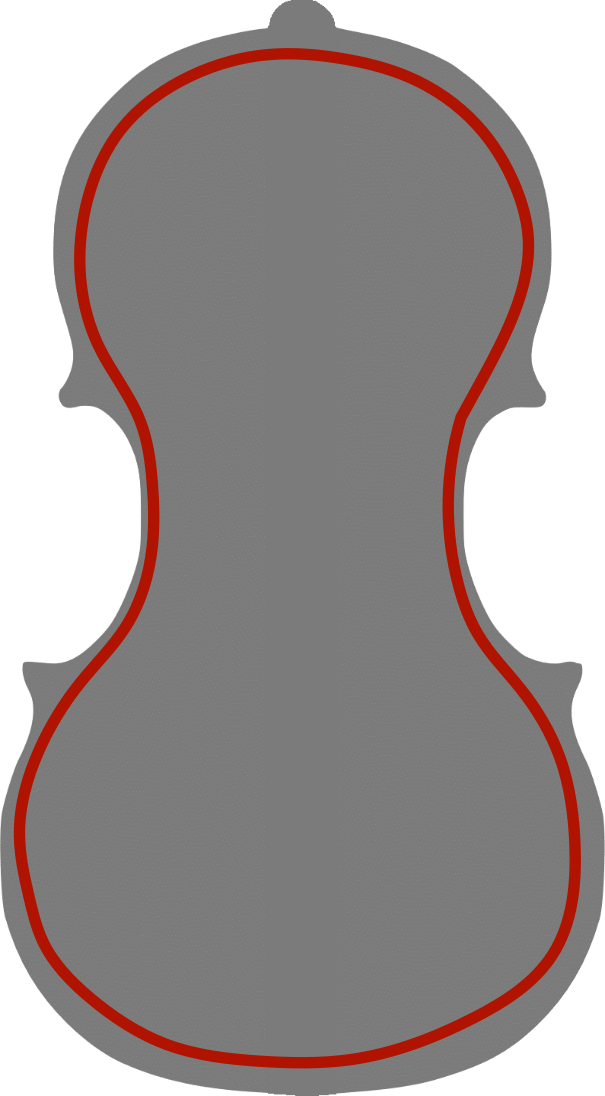}
  \end{subfigure}%
  \hspace{-0.5cm}% Space between image A and B
  \begin{subfigure}{0.24\linewidth}
    \centering
    \includegraphics[scale = 0.13]{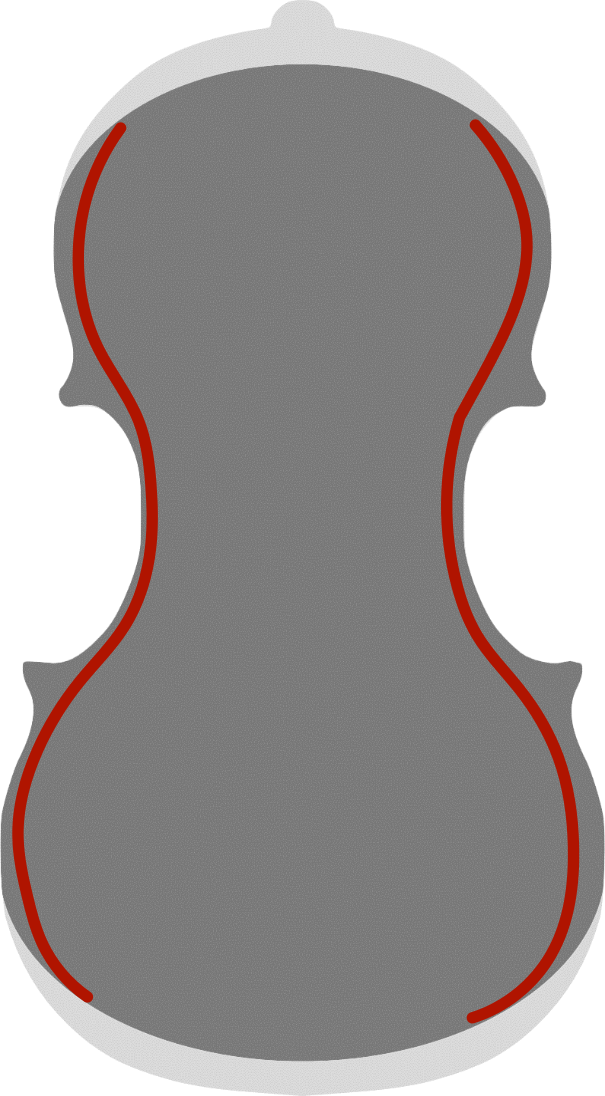}
  \end{subfigure}%
  \hspace{0.5cm}% Space between image B and C
  \begin{subfigure}{0.27\linewidth}
    \centering
    \includegraphics[scale = 0.13]{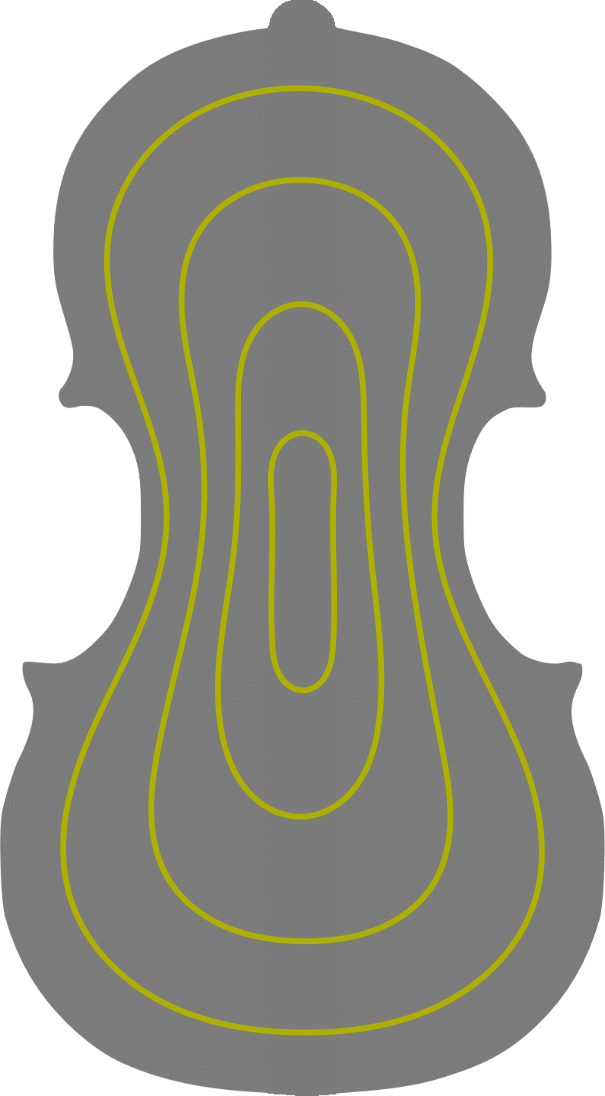}
  \end{subfigure}
  \hspace{-0.5cm}% Space between image C and D
  \begin{subfigure}{0.24\linewidth}
    \centering
    \includegraphics[scale = 0.13]{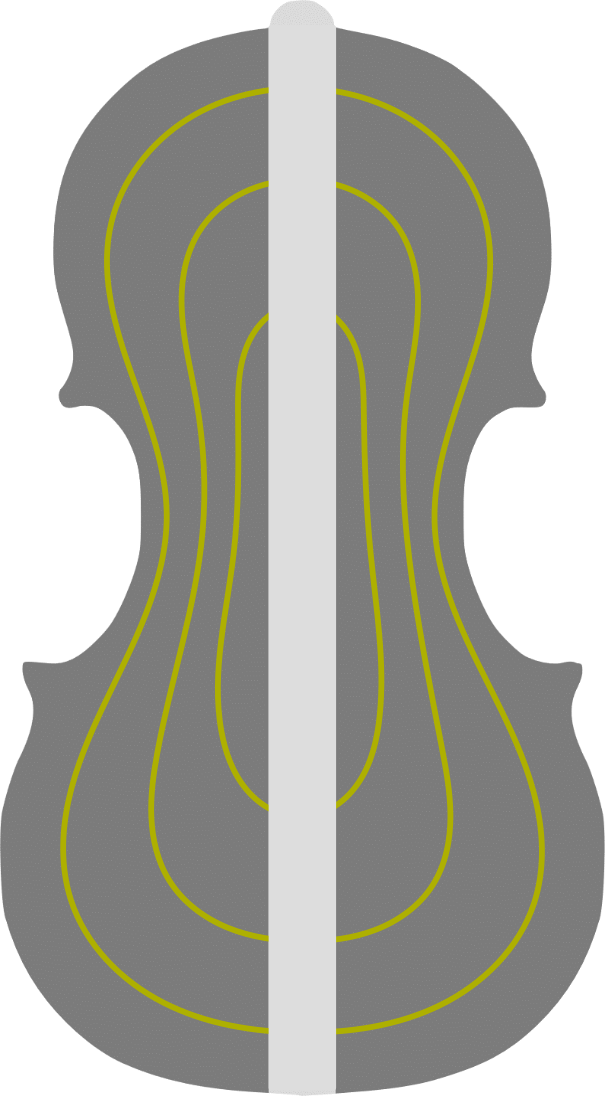}
  \end{subfigure} 
  \vspace{-0.2cm}
  \caption{Impact of the reduction of the length of the sound box on the channel of minima (left) and impact of the reduction of the width of the sound box on the contour lines (right)}
  \label{fig:Two_reductions}
\end{figure}

\vspace{-0.2 cm}\noindent Historical testimonies mention two main types of reduction, illustrated in Fig. \ref{fig:Two_reductions}. The first one consists of removing a crescent of wood from the top and bottom of the sound box to reduce the length of the instrument. The second involves cutting a slice of wood from the main axis of the instrument and gluing the two parts back together, to reduce the width of the violin. A more detailed historical explanation can be found in \cite{ceulemans2023baroque}. Recall that, in the violin family, the sound board and the back are not flat, but carved in a three-dimensional shape, often named arching. The above two kinds of reduction have an impact on two geometric characteristics of violins, namely the channel of minima and the contour lines. The channel of minima, also named fluting and highlighted in red in Fig. \ref{fig:Two_reductions}, is a groove carved into the sound board and back of a violin and running around these plates. When material is removed from the top and bottom of the sound box, this channel tends to disappear at both ends of the plates, as shown on the left of Fig. \ref{fig:Two_reductions}. On the right of Fig. \ref{fig:Two_reductions}, the yellow contour lines of the violin show smooth curves all along. When we remove a piece of wood from the centre and glue the two parts back together, the contour lines will become much sharper and more discontinuous along the central axis. 
\vspace{-0.5cm}

\noindent We have recently observed such geometric differences for both characteristics in a premilinary work involving only two instruments (one reduced and one unreduced) \cite{beghin2023validation}. Despite the frequency with which reductions were made to the sound boxes of violins, it is surprising that studies devoted to this practice are scarce. One study quantified the width of the slice of wood removed, typical of the second type of reduction, for two instruments by Andrea Amati that bore a painted coat of arms. After reconstructing the original coat of arms and determining the reduction it has undergone, it was then possible to assess to the tenth of millimetre how much wood had been removed \cite{radepont2020revealing}. A previous article only mentioned the reduction of one of these two Amati's instruments \cite{houssay2015cordes}. To our knowledge, this is the only other literature on violin reduction available to date. \\ 

\noindent It should be noted, however, that although almost no studies have been carried out on the reduction of violins, many more have focused on the creation of 3D models for other types of measurement. Numerous acquisition techniques include laser scanning 
\cite{dondi2016measuring,dondi20173d,fiocco2021compositional,fioravanti2012structural}, medical and industrial X-ray computed tomography (CT) scans \cite{frohlich2009secrets,fuchs2018musices,kirsch2015x,lothairecharacterization,marschke2018modeling,marschke2020approach,plath2017post,plath20193d,pyrkosz2013comparative,pyrkosz2011converting,pyrkosz2014coupled,stanciu2021x}, UV fluorescence with the use of a Kinect device \cite{dondi2018multimodal}, neutron imaging \cite{kirsch2015x} and finally, photogrammetry \cite{beghindigital,motteconception,pinto2008photogrammetric}. We chose to use photogrammetry to create our 3D models of the instruments.\\

\noindent First, in Section \ref{sec:photogrammetric meshes}, we describe the data we are working on, namely photogrammetric triangular meshes obtained thanks to photogrammetry. Then, in Section \ref{sec:Plane of Symmetry}, we focus on the notion of a plane of symmetry between the sound board and the back of an instrument. This notion will enable us to characterise and study contour lines and the channel of minima, described in Sections \ref{sec:Contour lines} and \ref{sec:Channel of minima} respectively. 

\section{Materials: Photogrammetric Meshes}
\label{sec:photogrammetric meshes}

\noindent Photogrammetry involves digitally recreating 3D objects based exclusively on 2D pictures. With about 150 photos per instrument and careful pre-processing, we are able to recreate models\footnote{We used the photogrammetry software Metashape, by Agisoft. See \url{https://www.agisoft.com/}.} of the outer surface that have been assessed and validated, and whose accuracy is similar to that of medical CT scans to within 0.2--0.3 \SI{}{\mm} \cite{beghin2023validation}. In this paper, we will show some examples of the differences between reduced and unreduced instruments, obtained from a corpus of 38 early violins, violas and cellos from the Musical Instruments Museum (MIM) in Brussels. The list with a few additional information can be found in Appendix \ref{sec:Appendix}. We had already shown examples for two violas (one reduced, the other not) for which the differences were clear \cite{beghin2023validation}. We will see in this paper that the full spectrum of geometric characteristics found in the larger corpus is actually more difficult to characterise. \\

\begin{figure}[H]
    \centering
    \begin{subfigure}{0.48\textwidth}
    \centering
    \includegraphics[height=3.8cm]{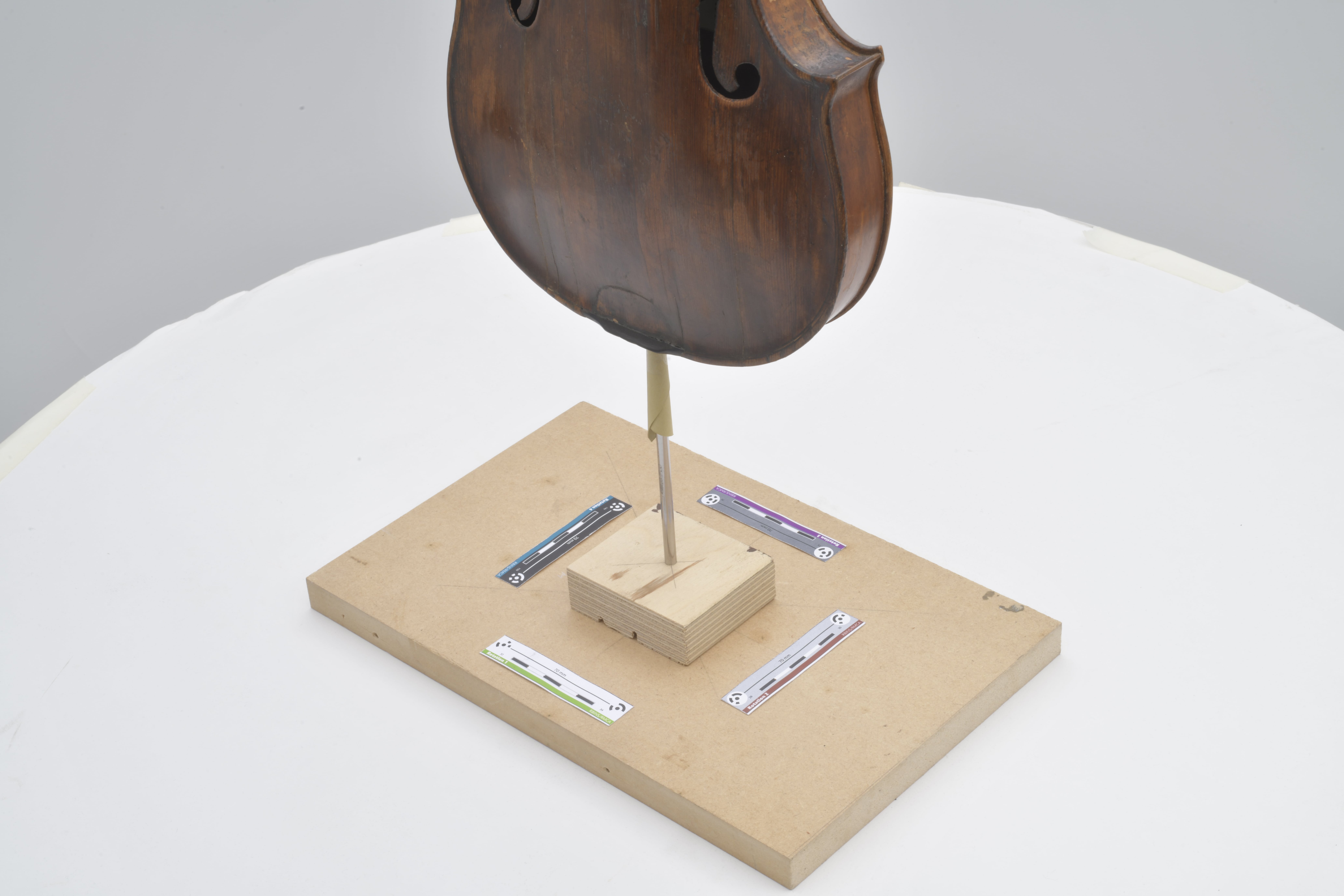}
    \end{subfigure}
    \hfill
    \begin{subfigure}{0.48\textwidth}
    \centering
    \includegraphics[height=3.8cm]{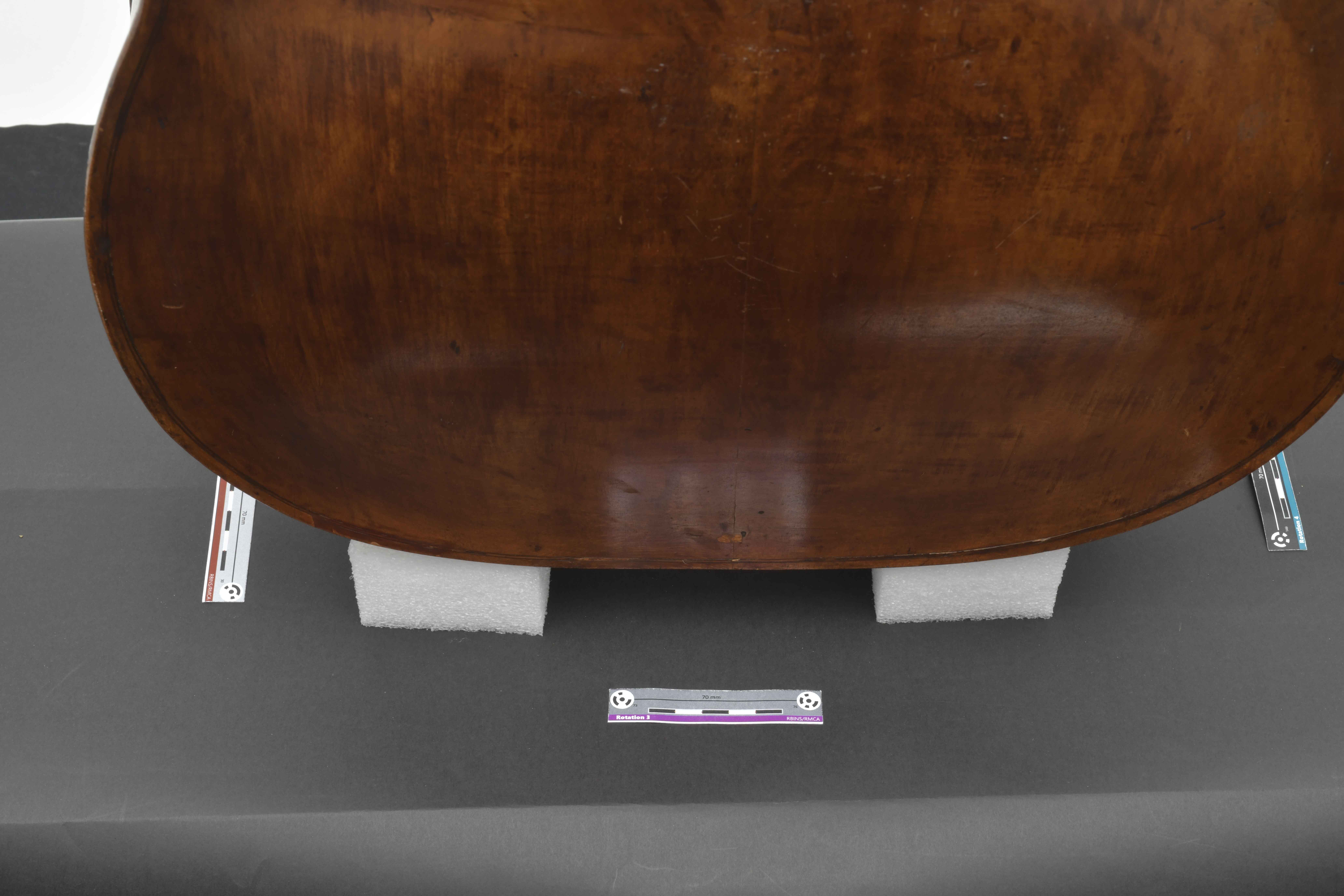}
    \end{subfigure}
    \caption{Four scale bars are used during the data acquisition process.\\ Examples for a violin (left) and a cello (right)}
    \label{fig:scale bar}
\end{figure}

\noindent One difference in acquiring the 36 additional instruments compared to the two we already had in \cite{beghin2023validation} is that scale bars were used in the photography process, as shown in Fig. \ref{fig:scale bar}. In this way, the meshes created are directly at the right scale and it is not necessary to resize them using a scaling factor. After the instruments have been fully acquired and digitally recreated, the necks are manually removed in the MeshLab\footnote{\url{https://www.meshlab.net/}} 3D editing software, leaving only the body of the instrument (consisting of the sound board, back and ribs). An initial alignment of the body is carried out using Principal Component Analysis (PCA) directly in MeshLab. The sound boards and backs are then simplified and automatically delineated. Ultimately, after the delineation process, the meshes of the plates are composed of about 80k nodes. The complete procedure is detailed in \cite{beghin2023validation}.
\vspace{-0.5cm}

\vspace{-1cm}
\section{Discussion on the Plane of Symmetry}
\label{sec:Plane of Symmetry}
\begin{figure}[H]
    \centering
    \begin{subfigure}{\textwidth}
    \centering
    \includegraphics[height = 3 cm]{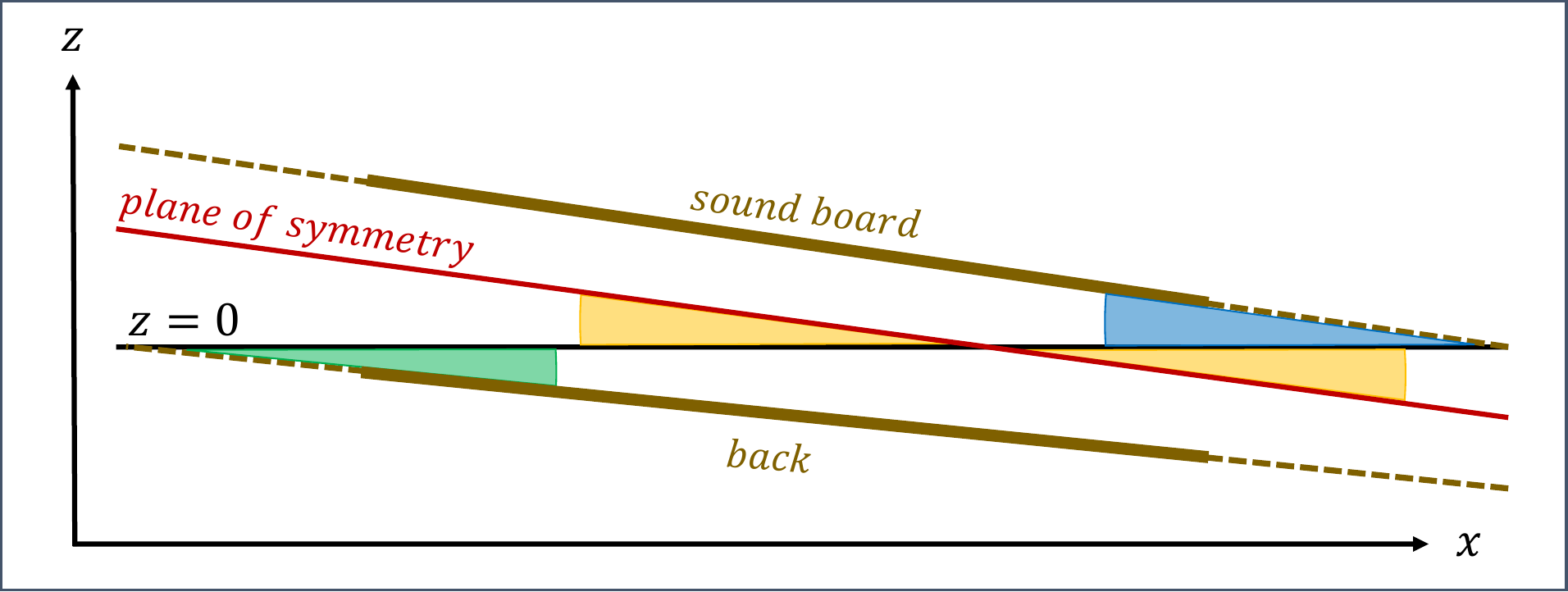}
    \end{subfigure}
    \hfill
    \begin{subfigure}{\textwidth}
    \centering
    \includegraphics[height = 3 cm]{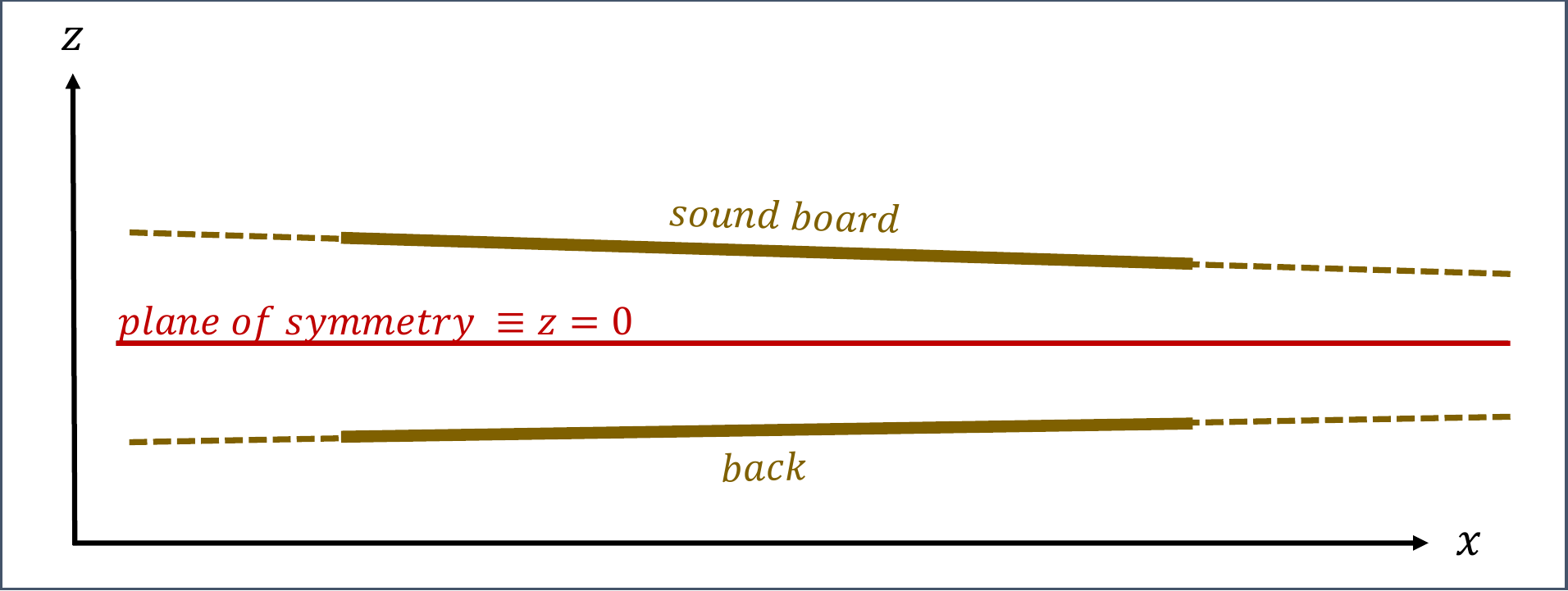}
    \end{subfigure}
    \vspace{-0.6cm}
    \caption{Planes of the sound board and back before (top) and after (bottom) the plane of symmetry has been matched with the plane $z=0$, i.e. after a 3D rotation (which does not correspond to a 2D rotation of the yellow angle) and the offset adjustment}
    \label{fig:symmetry_plane}
\end{figure}

As wood is a material that ages, warps and twists over time, it is unlikely to find a violin with perfectly aligned and symmetrical sound board and back. Still, we are interested in finding the plane of symmetry that best represents this expectation of parallelism between the plates. To do this, we first isolate the contours of the sound board and the back. Using orthogonal regression, we fit each of these two contours in a plane according to a least-square criterion. An example of a cross section in the $Oxz$ plane is shown in Fig. \ref{fig:symmetry_plane} (top). We then calculate the bisector plane between the planes of the two plates, namely the plane of symmetry between the sound board and the back, shown in red in Fig. \ref{fig:symmetry_plane}. Once this plane has been calculated, we apply a rotation operator to the coordinates of the plates, so that the plane of symmetry becomes parallel to the plane $z=0$. We finally adjust its offset to merge it with the $Oxy$ plane. The resulting alignment in the $Oxz$ plane is shown at the bottom of Fig. \ref{fig:symmetry_plane}. Note that as the example in Fig. \ref{fig:symmetry_plane} stems from a 2D planar cut and that the angle of rotation is a dihedral angle in 3D, the 3D rotation itself does not correspond to the yellow angle. 

\vspace{-0.5cm}
\begin{figure}[H]
\centering
\begin{subfigure}[b]{0.475\textwidth}
\hspace*{-0.7cm}\includegraphics[height=4.2cm]{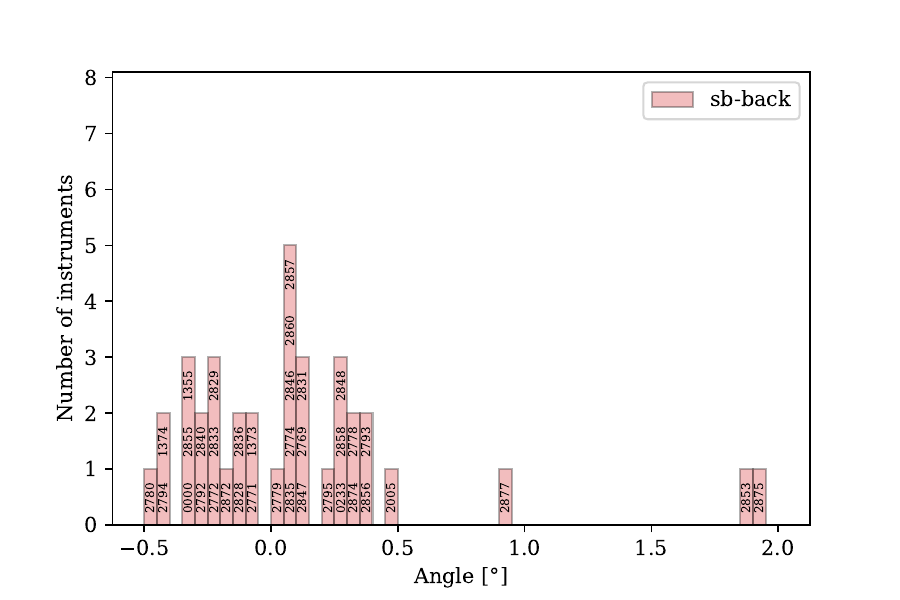}
\vspace{\baselineskip}
\hspace*{-0.7cm}\includegraphics[height=4.2cm]{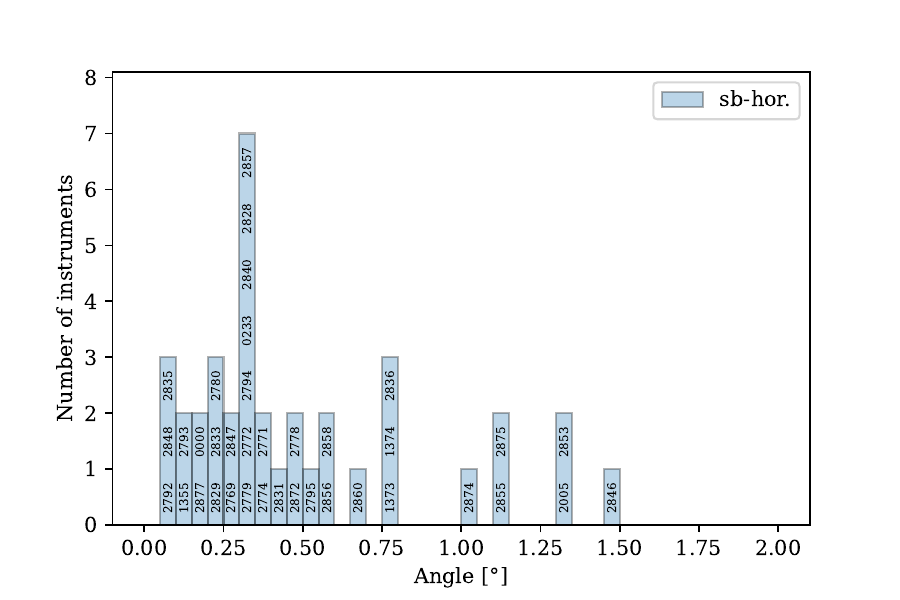}
\end{subfigure}%
\begin{subfigure}[b]{0.475\textwidth}
\includegraphics[height=4.2cm]{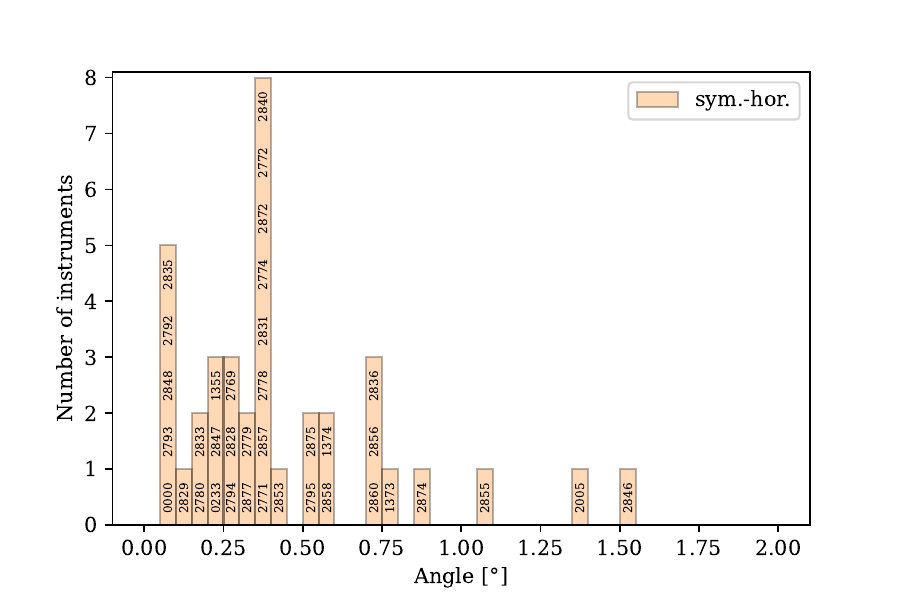}
\vspace{\baselineskip}
\includegraphics[height=4.2cm]{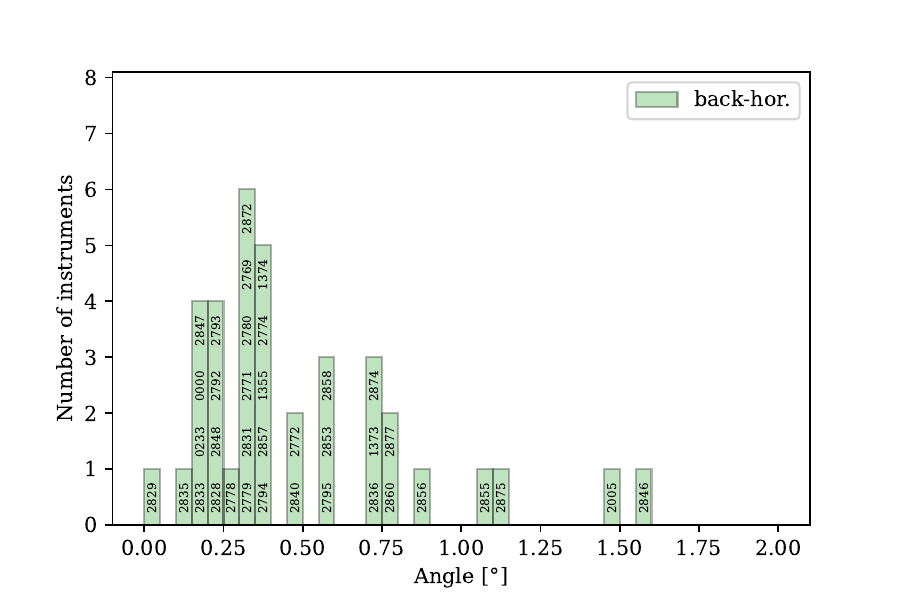}
\end{subfigure}%
\vspace{-0.7cm}
\caption{Distribution of the dihedral angles between the planes of the sound board and the back (top-left, red), the plane of symmetry and the horizontal plane (top-right, yellow), the sound board and the horizontal plane (bottom-left, blue), and the back and the horizontal plane (bottom-right, green)}
\label{fig:Angles}
\end{figure}

\noindent We were then interested in the different dihedral angles occuring between the planes of the contours of the sound board, the back, the plane of symmetry (before rotation) and the horizontal plane $z=0$. The distribution of these angles for the corpus of 37 instruments\footnote{Instrument 2776 has been disassembled and no longer has a sound board, only a back. Analysis of the various angles therefore makes no sense for this instrument.} is shown in Fig. \ref{fig:Angles}. On the histograms, bins are drawn every $\ang{0.05}$ and each instrument is referenced by its MIM inventory number\footnote{For the sake of convenience, instrument 2005.023 has been listed as 2005 on Fig. \ref{fig:Angles}.}, available in Appendix \ref{sec:Appendix}.

\noindent The angle of the plane between the sound board and the back (Fig. \ref{fig:Angles}, red, top-left) indicates the degree of parallelism of the plates. It is defined as positive if the sound board and back are closer at the bottom of the body than at the neck, which is the case for 21 out of 37 instruments, and negative if the sound board and the back are closer together at the back. These angles are generally less than $\ang{1}$, indicating near parallelism between the plates. The two steepest angles (for instruments 2853 and 2875) refer to a tenor violin and a cello whose bodies are completely twisted, which explains why the sound board and back are not well aligned. The lowest angle values (in absolute value) are mostly related to violins and violas, while the highest ones are linked to cellos. However, in the middle of the range of those parallelism angles, we find both violins and cellos.\\

\noindent The angle between the plane of symmetry and the horizontal plane, before they merged as a result of the 3D rotation and the offset adjustment (Fig. \ref{fig:Angles}, yellow, top-right) indicates the (non-)alignment of the violin body (sound board, back and ribs) with the axes of the PCA mentioned in Section \ref{sec:photogrammetric meshes}. We recall that the PCA is an alignment procedure based on the whole mesh of the body, and not only the two plates. All the angles larger than $\ang{0.5}$ concern only cellos, with the exception of 2795 and 2836 (which are very close to $\ang{0.5}$) and 2846, which has the greatest deviation from the horizontal plane, with an angle of more than $\ang{1.5}$. The higher values for the cellos indicate a less good alignment with the axes of the PCA than violins and violas. This can be explained by the fact that in a cello body mesh, the ribs are higher than in a violin or viola mesh. They therefore tend to influence the PCA more in their favour rather than in favour of the plates. \\

\noindent Interestingly, the instruments with the largest angle between the sound board and back are not the ones with the greatest angle between the plane of symmetry and the horizontal plane. For example, instruments 2853 and 2875, which have the largest angle between the sound board and the back (remember that they are twisted), are not the least well aligned with the initial PCA. On the other hand, we note that 2005.023, which is the least well aligned with the initial PCA, has a sound board and a back that are relatively well aligned. \\

\noindent Finally, the angles between the sound board and the horizontal plane (Fig. \ref{fig:Angles}, blue, bottom-left), and between the back and the horizontal plane (Fig. \ref{fig:Angles}, green, bottom-right) are given as a complement of information.

\section{Contour Lines}
\label{sec:Contour lines}

\begin{figure}[H]
    \centering
\includegraphics[scale=0.35]{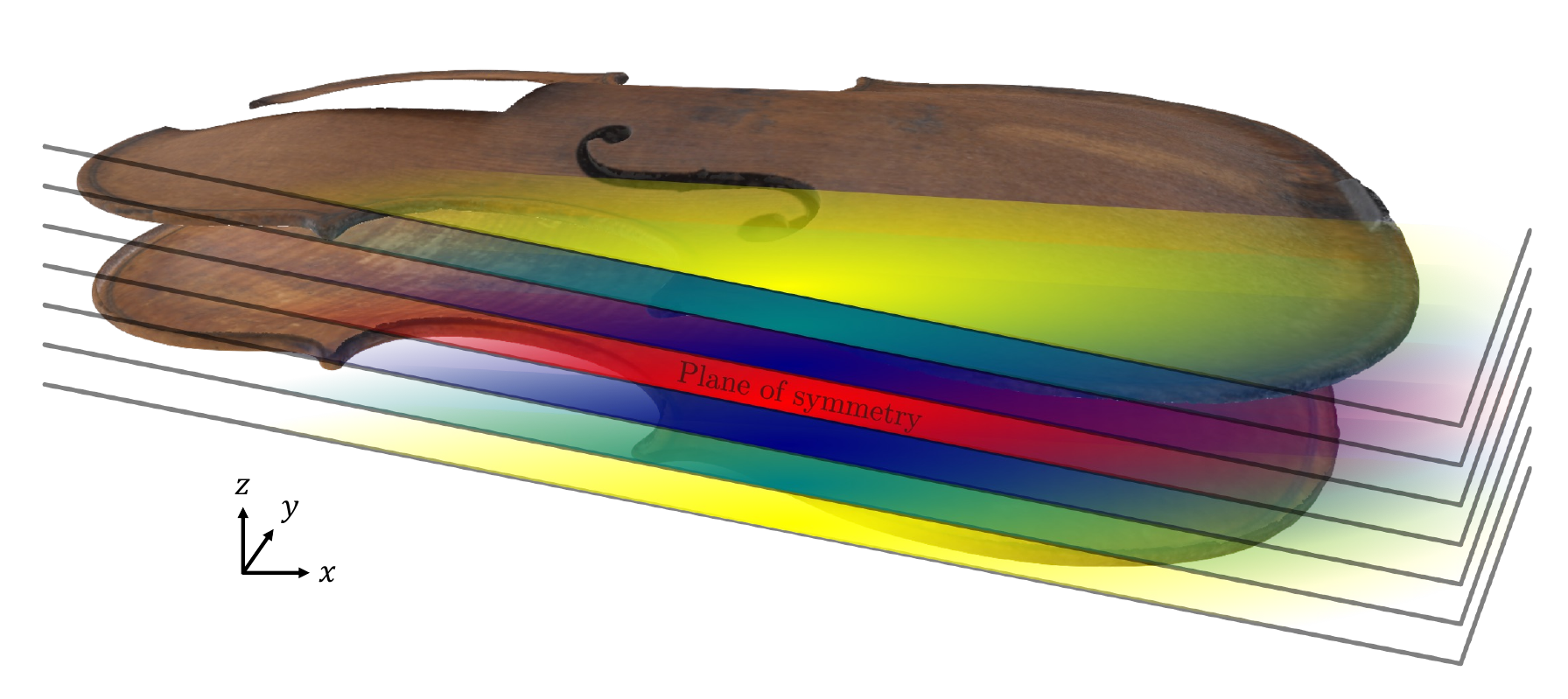}
    \caption{Plane of symmetry and successive parallel horizontal planes}
    \label{fig:g6080}
\end{figure}

\begin{figure}[H]
\centering
\begin{subfigure}[b]{0.475\textwidth}
\includegraphics[scale = 0.4]{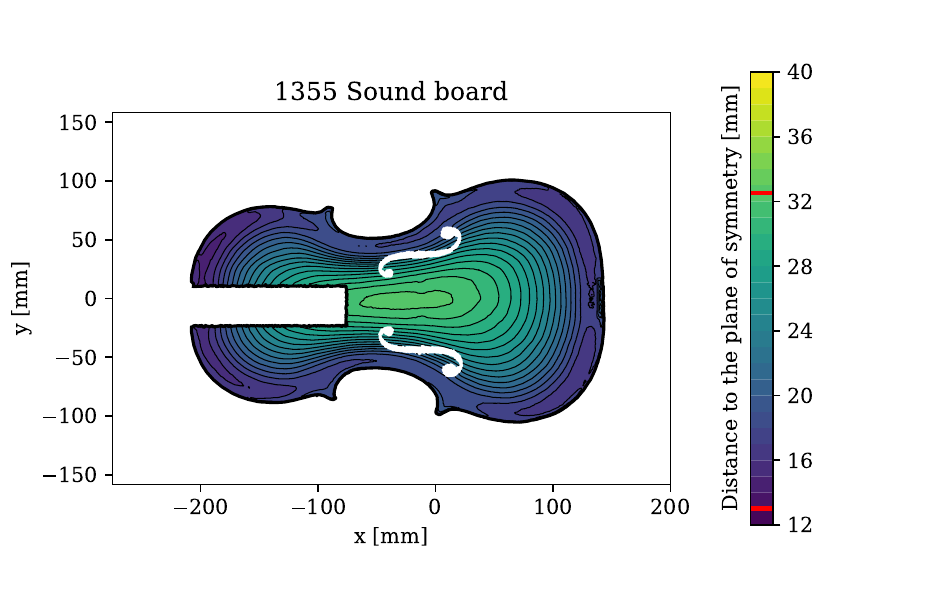}
\vspace{\baselineskip}
\includegraphics[scale = 0.4]{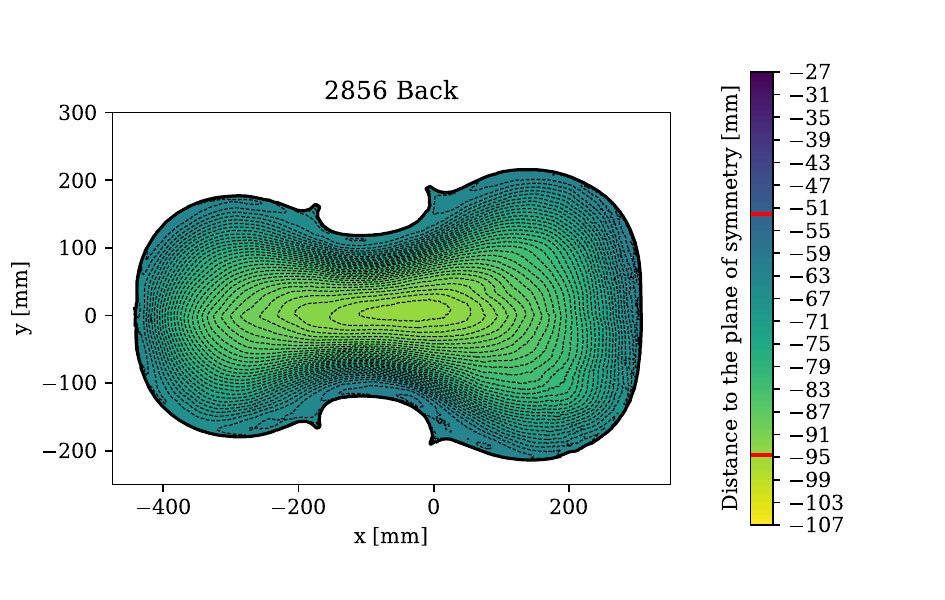}
\end{subfigure}%
\begin{subfigure}[b]{0.475\textwidth}
\includegraphics[scale = 0.4]{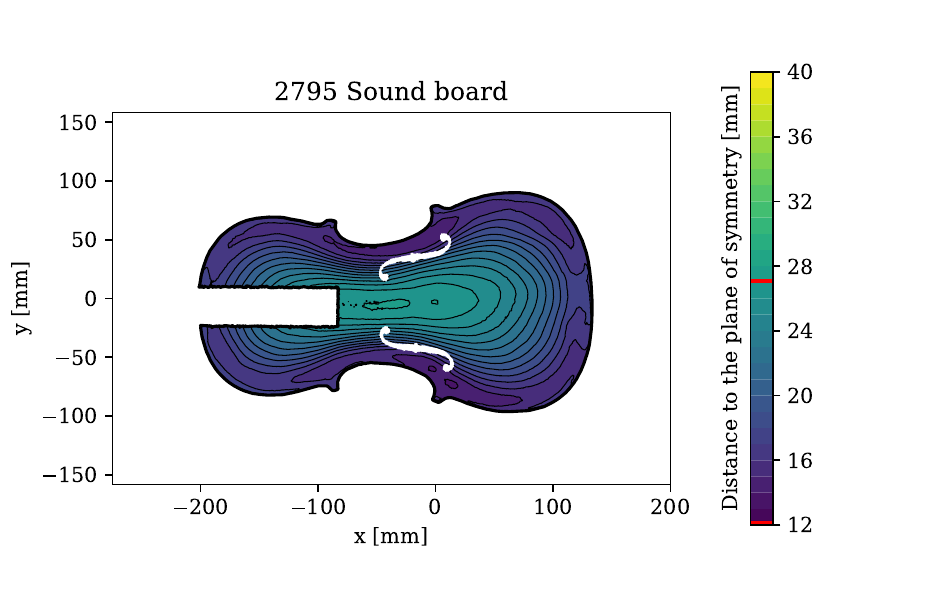}
\vspace{\baselineskip}
\includegraphics[scale = 0.4]{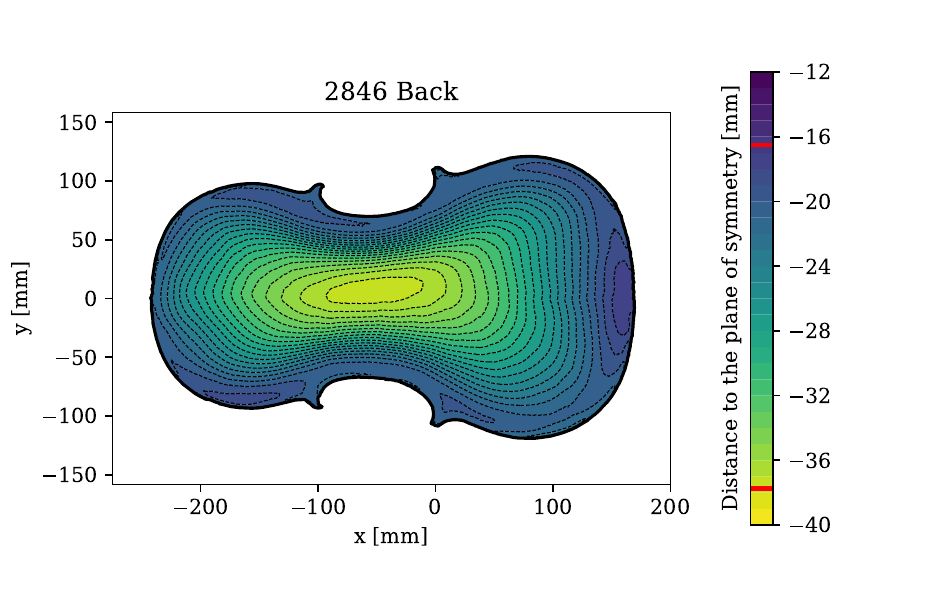}
\end{subfigure}%
\caption{Contour lines of four instruments' plates, drawn every \SI{}{\mm}: two violins (top), a cello (bottom-left) and a viola (bottom-right)}
\label{fig:contour_lines}
\end{figure}

\noindent Once the plane of symmetry has been identified, oriented horizontally and merged with the plane $z=0$, we can use it as a reference to calculate contour lines. As shown on Fig. \ref{fig:g6080}, we draw horizontal planes, parallel to the plane of symmetry, every millimetre in the direction of the sound board and the back, and study the resulting curves obtained as the intersections with the mesh. The results for four instruments in the corpus are shown in Fig. \ref{fig:contour_lines}. In this figure, two scales have been used: one for violins and violas, and the other for cellos. The colour bar on the right has a range of $\pm$ \SI{28}{\mm} for violins and violas and $\pm$ \SI{80}{\mm} for cellos (positive distance for sound boards, negative for backs). Such a high range for cellos is explained by the presence of instrument 2853 in the corpus. More information is given in the Appendix \ref{sec:Appendix}. Finally, the red interval corresponds to the exact range of elevation of the plate with respect to the plane of symmetry.\footnote{We have therefore slightly changed the colour bar convention used in \cite{beghin2023validation} to emphasise the distance from the plane of symmetry, rather than the height of the sound boards or the depth of the backs. This information can also be seen in red on the colour bars.} 

\noindent We see on the 1355 violin sound board (top-left in Fig. \ref{fig:contour_lines}) very round and circular contour lines, typical of an instrument at the time of its creation, which has not been reduced. For the violin 2795, we see that those contour lines tend to be a little sharper around the axis of symmetry, and that a plateau appears at the end of the sound board, indicating a possible reduction. The suspicions of reduction are even more pronounced for the two lower instruments, the 2856 cello and 2846 viola. In both cases, we see that the curves at the top of the back are very sharp. Interestingly, in the case of the 2846 viola, the curves at the bottom of the back are completely flat, which draws our attention to a possible reduction as well, or at least a modification of the instrument. Indeed, we have found empirically that this phenomenon can also occur on reduced violins, after they have been dismantled and reassembled with difficulty.\\

\noindent In our previous article \cite{beghin2023validation}, we worked on two violas with very different contour lines. Our results and discussions with experts clearly allowed us to conclude whether or not the instruments had been reduced. With an extended corpus, the interpretation of some of the 38 instruments becomes more difficult. Since the spectrum of geometric characteristics is much more varied, the information is sometimes harder to analyse.

\section{Channel of Minima}
\label{sec:Channel of minima}

\noindent The channel of minima, already defined in Section \ref{sec:Introduction} and highlighted in red in Fig. \ref{fig:Two_reductions}, is a groove carved into the sound board or back of a violin and running around these plates. We propose a new approach to calculate this channel, which is intended to be faster, more robust and more faithful to organological reality than the one we developed in \cite{beghin2023validation}. In that preliminary study, we were performing planar cuts orthogonal to the contour of the plates and to the plane of symmetry, then approximated the intersection curve (corresponding to the surface of the plate) by a spline and extracted a global minimum from each of these `spline cuts'. This procedure was applied to a planar cut every millimetre, for the left and right sides of each plate. The planar cuts were made using the Python package \texttt{meshcut}\footnote{\url{https://github.com/julienr/meshcut}}, which is not computationally efficient. \\

\begin{figure}[H]
\centering
\begin{subfigure}[b]{0.475\textwidth}
\includegraphics[height = 4 cm]{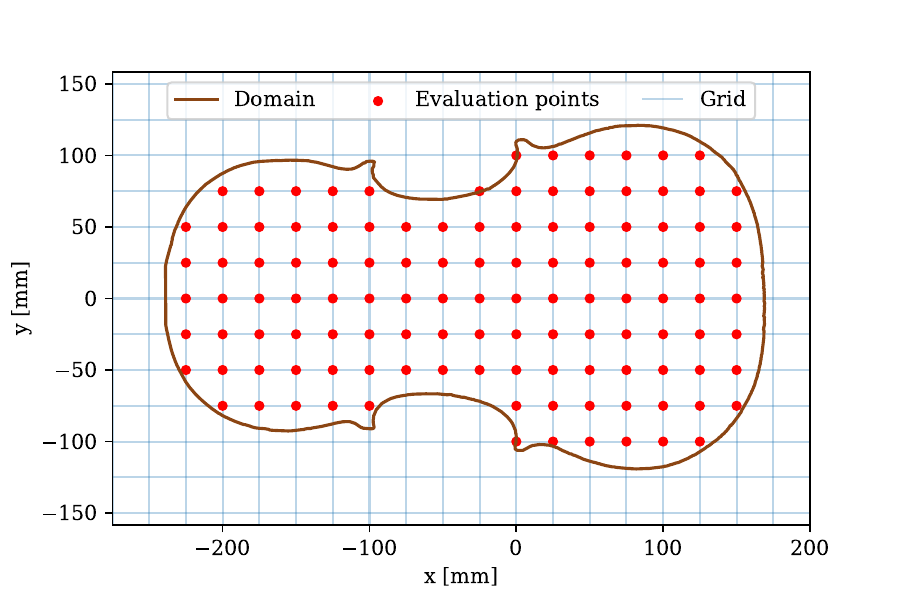}
\vspace{\baselineskip}
\includegraphics[height = 4 cm]{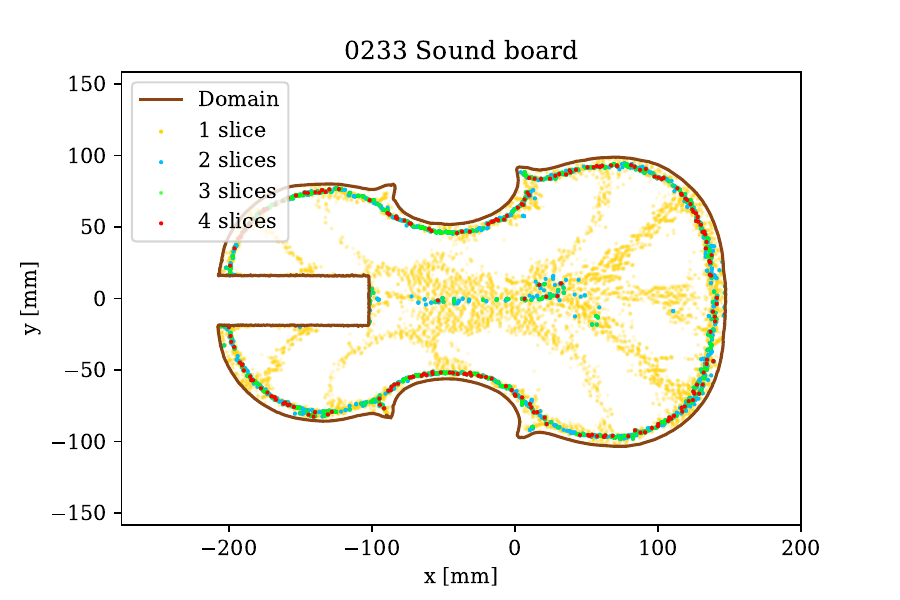}
\end{subfigure}%
\begin{subfigure}[b]{0.475\textwidth}
\includegraphics[height = 4 cm]{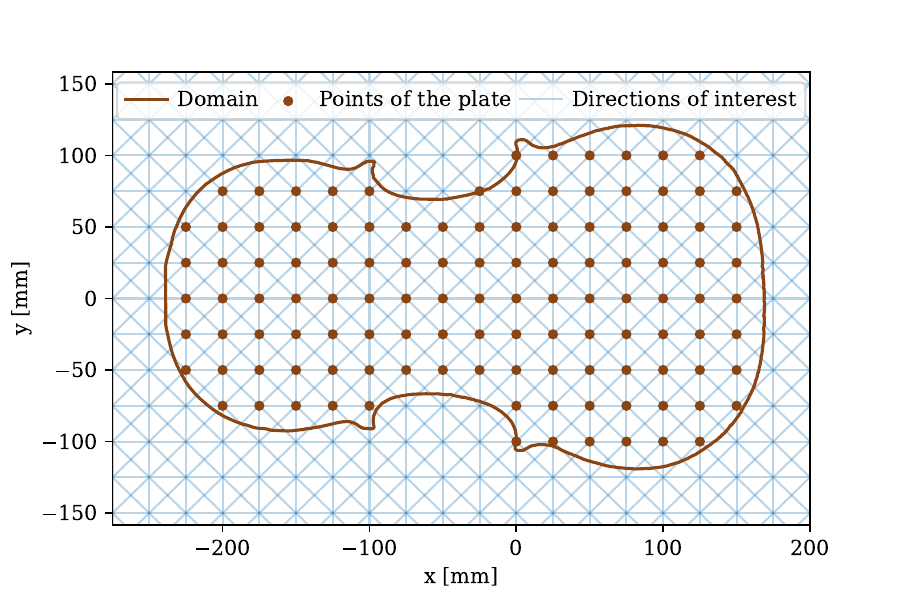}
\vspace{\baselineskip}
\includegraphics[height = 4 cm]{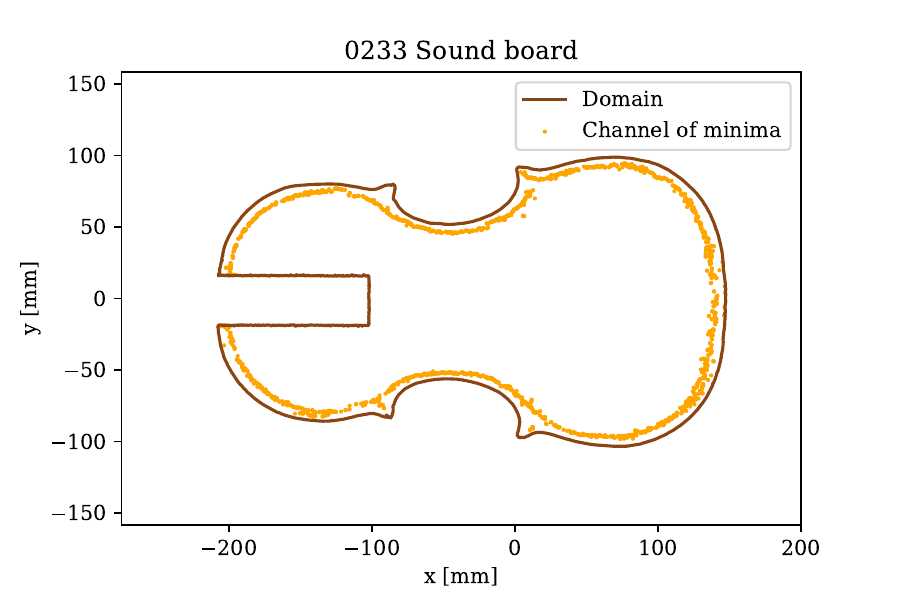}
\end{subfigure}%
\vspace{-0.5 cm} 
\caption{Linear sampling of the mesh over a regular grid (top-left) and computation of the channel of minima (top-right). Local minima on slices (bottom-left) and resulting channel after filtering (bottom-right) for the 0233 violin.}
\label{fig:channel_minima_theory}
\end{figure}

\noindent Instead, we now sample each sound board and back over a horizontal grid, refined every \SI{0.25}{\mm} in each direction and parallel to the plane of symmetry (after rotation) of the plates. We thus move from a surface mesh with disordered vertices to a map elevation with ordered and regular vertices. This is illustrated in Fig. \ref{fig:channel_minima_theory} (top-left), albeit with much less refinement than the actual computations. Next, we look at sets of points that are aligned either horizontally, vertically or diagonally (both in $\pm 45 \degree$ directions) and keep the local minima on each of these `slices', as in Fig. \ref{fig:channel_minima_theory} (top-right), where the blue lines indicate the considered alignments. A minimum is considered to be local if it is smaller than all its neighbours within a distance of \SI{2}{\mm} for violins and violas and \SI{5}{\mm} for cellos\footnote{In the neighbourhood of the domain, we only take into account the available nodes. For example, if the second node of a `slice' is lower than the first node (which corresponds to the edge of the instrument in a direction) and all the others in the other direction, within a distance of \SI{2}{\mm} or \SI{5}{\mm}, then this second node is considered as a local minimum.}. Finally, we consider that a point belongs to the channel of minima if it is a local minimum in at least two of the four `slices' through which it passes. This approach seems more reliable than the one used in \cite{beghin2023validation}, as it gives real meaning to the notion of a channel carved in three dimensions (and therefore in several directions, and not only along a single plane). As the planar cuts used previously were approximated by splines, wrong estimates on the cuts could return false global minima. Furthermore, a finite set of points (in this case, a spline estimated every \SI{}{\mm}) systematically possessed a global minimum, but might not correspond to the carved intuition behind the channel of minima, as in our new approach. We can see in Fig. \ref{fig:channel_minima_theory} (bottom-left) the minima that are present respectively on one, two, three and four of the slices, and we finally show the resulting channel on bottom right. In addition to keeping only the points common to at least two slices, we filter out the outliers located on the archings of the plates.
\vspace{-2 cm} 
\begin{figure}[H]
\centering
\begin{subfigure}[b]{0.475\textwidth}
\includegraphics[scale = 0.4]{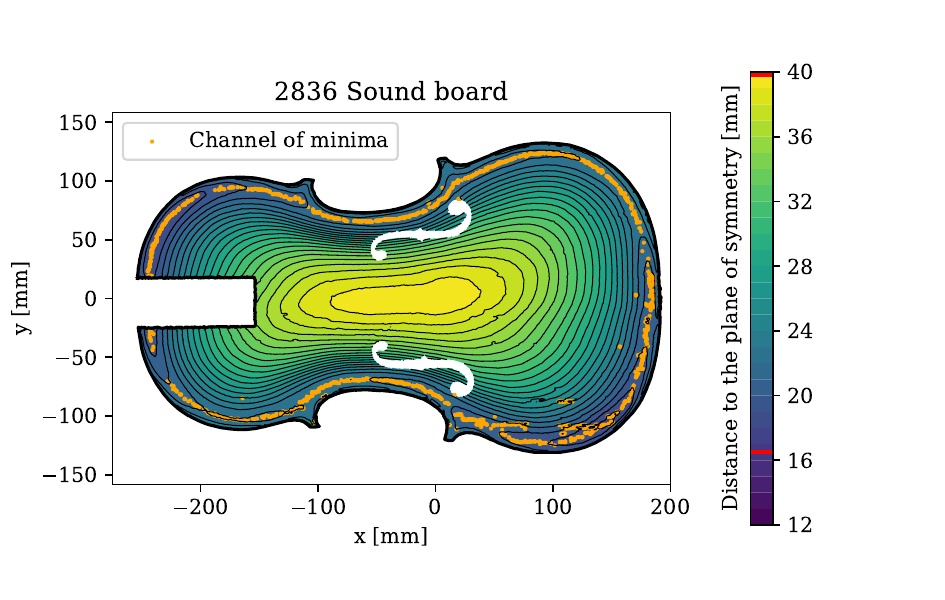}
\vspace{\baselineskip}
\includegraphics[scale = 0.4]{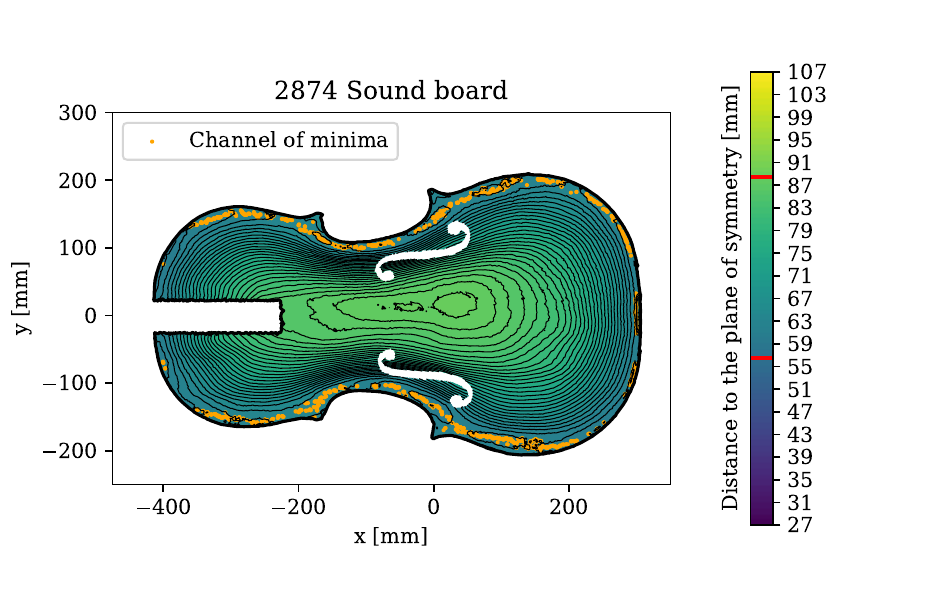}
\end{subfigure}%
\begin{subfigure}[b]{0.475\textwidth}
\includegraphics[scale = 0.4]{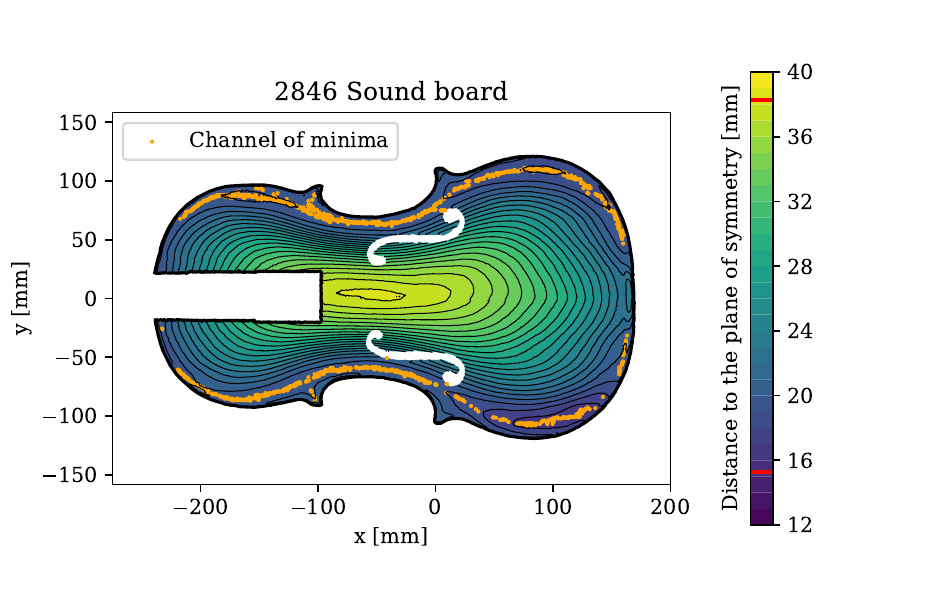}
\vspace{\baselineskip}
\includegraphics[scale = 0.4]{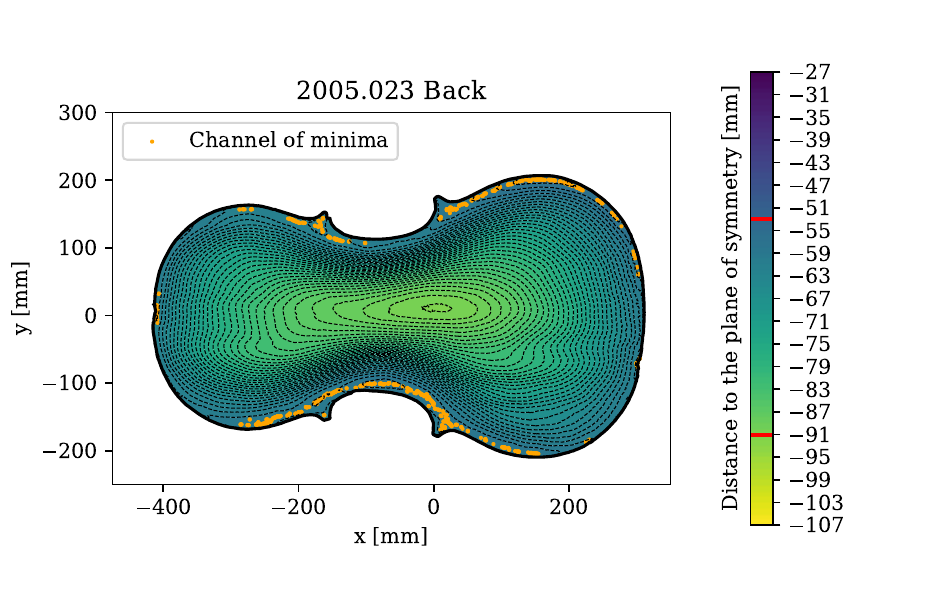}
\end{subfigure}%
\caption{Channel of minima (displayed in orange) with contour lines of four instruments' plates, drawn every \SI{}{\mm}: two violas\protect\footnotemark (top) and two cellos (bottom)}
\label{fig:channel_minima_practical}
\end{figure}

\footnotetext{The museum denomination of instrument 2836 is a tenor violin, but it is associated to violas. More about it can be found in the Appendix \ref{sec:Appendix}.}

\noindent The results for four instruments are highlighted in Fig. \ref{fig:channel_minima_practical}. The local minima common to at least two `slices' are presented in raw form (as orange dots). We could have interpolated them for a clearer rendering, but we preferred to leave them in order to highlight the trends of the channel. We have also illustrated the contour lines to show their link with the channel. The colour bar on the right should be read in exactly the same way as explained in Section \ref{sec:Contour lines} and Fig. \ref{fig:contour_lines}. \\

\noindent The 2836 viola in the top left has a channel that runs continuously through the whole sound board, whereas the other three have a discontinuous path, or even almost none at all. By comparing these results with the historical data, we can highlight several points:

\begin{itemize}
    \item First of all, the attribution of viola 2846 (top-right) is not very clear. After discussions with luthiers and musicologists, it would appear to come from the same violin maker as 2836 (top-left). Furthermore, the MIM is very confident on the fact that instrument 2836 has been preserved in its original state, while 2846 has been reduced. It is therefore very likely that viola 2846 resembled 2836 before reduction. This interpretation is consistent with our results. 
    \item Then, we have very little information on cello 2874 (bottom-left). Nevertheless, it bears a label saying: "Recoupé à Tournay / par Ambroise De Comble 1782"(\textit{lit.} "Reduced in Tournai by Ambroise De Comble 1782"), which is an important indication in our research. Our results clearly match that label. 
    \item Before being reduced to the size of a cello, instrument 2005.023 (bottom-right) is thought to have been a bass violin. This would explain the poor outfit of our graph, although the channel of minima does tend to appear slightly on either side of the back. 
\end{itemize}

\noindent We add a comment on the contour lines of instrument 2836. We see that they are round, but slightly askew. This indicates that the instrument is twisted. Generally speaking, we realised in this research that the shape of the contour lines is not only an indicator of whether or not violins have been reduced, but that they also indicate that the instruments may have undergone other transformations such as twisting, flattening, a replacement of the sound board or the back, and sometimes even that the wood has cracked. The wide range of instruments present in our corpus makes the interpretation of curves more complex, both for the channel of minima and contour lines. 

\section{Conclusion}
\label{sec:Conclusion}

This research is the continuation of the preliminary work that we initiated in \cite{beghin2023validation}. We have made improvements to our data acquisition (using scale bars for photogrammetry) and to the calculation of the minima channel, with a faster, more robust and more interpretable approach. We have gone from an analysis of two violas to a corpus of 38 violins, violas and cellos. With this corpus, however, comes a spectrum of results that is not always easy to analyse, with sometimes contradictory information (historical data, empirical analysis by a luthier, contour lines, channel of minima, etc.). It is all part of the complexity of our research. In the end, we realise that in many cases it is not straightforward to determine whether an instrument has been reduced or not, but we can now certainly see that they have undergone transformations or deformations, have been twisted, bent, sometimes even cracked, etc. The next step in our research will be to focus on a more quantitative approach to the analysis of the curves (studies of contour lines and channel of minima with parametric functions, use of classification techniques from machine learning, etc.) to see whether or not these analyses can corroborate our qualitative analysis, presented in this article.

\appendix 

\section{List of Instruments in the Corpus}
\label{sec:Appendix}

This table lists the instruments studied in this research, classified by their inventory number at the Musical Instruments Museum (MIM), Brussels. The hyperlink on each number refers to the official MIM documentation and provides additional information. The question marks in the table indicate missing data or uncertainties. Finally, four instruments stand out slightly from the rest of the corpus: two tenor violins and two bass violins. These are names and sizes used before the standardisation of instruments, around 1750. The tenor violins are slightly larger than violas, and the bass violins are slightly larger than cellos.  For ease of reference, the 2836 has been compared with the same scale as violins and violas in Sections \ref{sec:Contour lines} and \ref{sec:Channel of minima}, while the 2853 and the two bass violins have been associated with cellos. It should be noted that although the two tenor violins share the same terminology, the two instruments are quite different. The 2853 has much larger ribs, and is thought to have been used as a child's cello, which explains its comparison with cellos (Table \ref{tab1}).

\begin{table}
\caption{List of instruments in the corpus}\label{tab1}
%\begin{tabular}{|l|l|l|}
\begin{tabular}{|p{1.2cm}|p{2.1cm}|p{5.5cm}|p{3cm}|}
\hline
Inv. Nr. & Size & Attribution & Date\\
\hline
\href{https://www.carmentis.be:443/eMP/eMuseumPlus?service=ExternalInterface&module=collection&objectId=109202&viewType=detailView}{0233} & Violin   & Matthys Hofmans IV    & Between 1645--1679\\
 \href{https://www.carmentis.be:443/eMP/eMuseumPlus?service=ExternalInterface&module=collection&objectId=109427&viewType=detailView}{1355} & Violin &   Matthys Hofmans IV  & Before 1672   \\
 
 \href{https://www.carmentis.be:443/eMP/eMuseumPlus?service=ExternalInterface&module=collection&objectId=109448&viewType=detailView}{1373} & Cello &   Marcus Snoeck  & 1718   \\

 \href{https://www.carmentis.be:443/eMP/eMuseumPlus?service=ExternalInterface&module=collection&objectId=109450&viewType=detailView}{1374} & Cello &   Gaspar Borbon  & 1688   \\

 \href{https://www.carmentis.be:443/eMP/eMuseumPlus?service=ExternalInterface&module=collection&objectId=110909&viewType=detailView}{2005.023} & Cello &   Gaspar Borbon  & 1707   \\
 \href{https://www.carmentis.be:443/eMP/eMuseumPlus?service=ExternalInterface&module=collection&objectId=109773&viewType=detailView}{2769} & Violin & Hendrick Willems & Between 1723--1743\\
 \href{https://www.carmentis.be:443/eMP/eMuseumPlus?service=ExternalInterface&module=collection&objectId=109775&viewType=detailView}{2771} & Violin    &Hendrick Willems & Between 1701--1730 \\
 \href{https://www.carmentis.be:443/eMP/eMuseumPlus?service=ExternalInterface&module=collection&objectId=109776&viewType=detailView}{2772} & Violin &   Hendrick Willems  & 1715 \\
\href{https://www.carmentis.be:443/eMP/eMuseumPlus?service=ExternalInterface&module=collection&objectId=109778&viewType=detailView}{2774} & Violin &   Gaspar Borbon  & Between 1650--1700   \\
\href{https://www.carmentis.be:443/eMP/eMuseumPlus?service=ExternalInterface&module=collection&objectId=109782&viewType=detailView}{2778} & Violin &   Egidius Snoeck  & Between 1700-1750   \\
\href{https://www.carmentis.be:443/eMP/eMuseumPlus?service=ExternalInterface&module=collection&objectId=109784&viewType=detailView}{2779} & Violin &   Egidius Snoeck  & 1727   \\
\href{https://www.carmentis.be:443/eMP/eMuseumPlus?service=ExternalInterface&module=collection&objectId=109785&viewType=detailView}{2780} & Violin &   Marcus Snoeck  & 1740   \\
\href{https://www.carmentis.be:443/eMP/eMuseumPlus?service=ExternalInterface&module=collection&objectId=109814&viewType=detailView}{2792} & Violin &   Matthys Hofmans IV  & 1665   \\
\href{https://www.carmentis.be:443/eMP/eMuseumPlus?service=ExternalInterface&module=collection&objectId=109815&viewType=detailView}{2793} & Violin &   M. Hofmans IV? G. Borbon?  & Before 1679   \\
\href{https://www.carmentis.be:443/eMP/eMuseumPlus?service=ExternalInterface&module=collection&objectId=109816&viewType=detailView}{2794} & Violin &   Jooris Willems  & 1659   \\
\href{https://www.carmentis.be:443/eMP/eMuseumPlus?service=ExternalInterface&module=collection&objectId=109817&viewType=detailView}{2795} & Violin &   Jooris Willems?  & 1659   \\
\href{https://www.carmentis.be:443/eMP/eMuseumPlus?service=ExternalInterface&module=collection&objectId=109882&viewType=detailView}{2828} & Viola &   Hendrick Willems?  & 1660   \\
\href{https://www.carmentis.be:443/eMP/eMuseumPlus?service=ExternalInterface&module=collection&objectId=109885&viewType=detailView}{2829} & Viola &   Hendrick Willems  & Between 1650--1659   \\
\href{https://www.carmentis.be:443/eMP/eMuseumPlus?service=ExternalInterface&module=collection&objectId=109923&viewType=detailView}{2831} & Viola &   Jooris Willems  & ?   \\
\href{https://www.carmentis.be:443/eMP/eMuseumPlus?service=ExternalInterface&module=collection&objectId=109925&viewType=detailView}{2833} & Viola &   Johannes Theodorus Cuypers?  & 1761   \\
\href{https://www.carmentis.be:443/eMP/eMuseumPlus?service=ExternalInterface&module=collection&objectId=109927&viewType=detailView}{2835} & Viola &   J. H. J. Rottenburgh? M. Snoeck?  & 1753   \\
\href{https://www.carmentis.be:443/eMP/eMuseumPlus?service=ExternalInterface&module=collection&objectId=109928&viewType=detailView}{2836} & Tenor Violin\tablefootnote{This tenor violin has been preserved in its original condition and has not been reduced. It is a typical example of an instrument whose original size was not standard, and which is slightly larger than a viola. } &   Gaspar Borbon  & 1692  \\
\href{https://www.carmentis.be:443/eMP/eMuseumPlus?service=ExternalInterface&module=collection&objectId=109932&viewType=detailView}{2840} & Viola &   Peeter I. Borbon? J.-J.-A. De Lannoy?  & 1771   \\
\href{https://www.carmentis.be:443/eMP/eMuseumPlus?service=ExternalInterface&module=collection&objectId=109939&viewType=detailView}{2846} & Viola &   M. Hofmans IV? G. Borbon?  & Before 1679   \\
\href{https://www.carmentis.be:443/eMP/eMuseumPlus?service=ExternalInterface&module=collection&objectId=109940&viewType=detailView}{2847} & Viola &   M. Hofmans IV? G. Borbon?  & Before 1691 \\
\href{https://www.carmentis.be:443/eMP/eMuseumPlus?service=ExternalInterface&module=collection&objectId=109941&viewType=detailView}{2848} & Viola &   Matthys Hofmans IV?  & Before 1672   \\

\href{https://www.carmentis.be:443/eMP/eMuseumPlus?service=ExternalInterface&module=collection&objectId=109946&viewType=detailView}{2853} & Tenor Violin &  Egidius Snoeck  & 1714 \\

\href{https://www.carmentis.be:443/eMP/eMuseumPlus?service=ExternalInterface&module=collection&objectId=109948&viewType=detailView}{2855} & Cello &  H. Willems? M. Snoeck? E. Snoeck?  & Between 1701-1750  \\

\href{https://www.carmentis.be:443/eMP/eMuseumPlus?service=ExternalInterface&module=collection&objectId=109949&viewType=detailView}{2856} & Cello &  Gaspar Borbon  & 1670  \\

\href{https://www.carmentis.be:443/eMP/eMuseumPlus?service=ExternalInterface&module=collection&objectId=109950&viewType=detailView}{2857} & Cello &  Gaspar Borbon  & 1671  \\

\href{https://www.carmentis.be:443/eMP/eMuseumPlus?service=ExternalInterface&module=collection&objectId=109951&viewType=detailView}{2858} & Cello &  Marcus Snoeck  & 1721  \\

\href{https://www.carmentis.be:443/eMP/eMuseumPlus?service=ExternalInterface&module=collection&objectId=109953&viewType=detailView}{2860} & Cello &  Marcus Snoeck  & 1733  \\

\href{https://www.carmentis.be:443/eMP/eMuseumPlus?service=ExternalInterface&module=collection&objectId=109965&viewType=detailView}{2872} & Cello & E. Snoeck? M. Snoeck?   &  1761 \\

\href{https://www.carmentis.be:443/eMP/eMuseumPlus?service=ExternalInterface&module=collection&objectId=109967&viewType=detailView}{2874}\tablefootnote{We have very little information on this instrument. Nevertheless, it bears a label saying: "Recoupé à Tournay / par Ambroise De Comble 1782"(\textit{lit.} "Reduced in Tournai by Ambroise De Comble 1782"), which is an important indication in our research.} & Cello &  ?  & Before 1782   \\

\href{https://www.carmentis.be:443/eMP/eMuseumPlus?service=ExternalInterface&module=collection&objectId=109968&viewType=detailView}{2875} & Bass Violin &  Egidius Snoeck  & 1734  \\

\href{https://www.carmentis.be:443/eMP/eMuseumPlus?service=ExternalInterface&module=collection&objectId=109970&viewType=detailView}{2877} & Bass Violin & Marcus Snoeck  & Before 1762   \\

0000\tablefootnote{This violin does not belong to the MIM. It is a recent instrument that we have acquired to be reduced by a luthier. This will enable us to compare its dimensions before and after reduction.} & Violin &   ?  & 2021   \\
\hline
\end{tabular}
\end{table}

\begin{comment}
 For citations of references, we prefer the use of square brackets
and consecutive numbers. Citations using labels or the author/year
convention are also acceptable. The following bibliography provides
a sample reference list with entries for journal
articles~\cite{ref_article1}, an LNCS chapter~\cite{ref_lncs1}, a
book~\cite{ref_book1}, proceedings without editors~\cite{ref_proc1},
and a homepage~\cite{ref_url1}. Multiple citations are grouped
\cite{ref_article1,ref_lncs1,ref_book1},
\cite{ref_article1,ref_book1,ref_proc1,ref_url1}.   
\end{comment}

\newpage

\begin{credits}
\subsubsection{\ackname} We thank Pr. Sandra Soares Frazao and her research team for sharing their best practices in photogrammetry. We would also like to thank Joris De Valck of the MIM (Brussels), Mirte Maes and Karel Moens of the Museum Vleeshuis (Antwerp) and Jan Strick of the Maison Bernard (Brussels). Their expertise on early instruments strengthened the analysis and discussion of our results. We thank Jean-Philippe Echard of the Musée de la Musique - Philharmonie de Paris for his careful reading and insightful comments on a previous version of this article. Finally, we would also like to thank Iona Thys for the photographic acquisition of the corpus of instruments. 

\subsubsection{\discintname}
The authors declare that they have no competing interests.
\end{credits}
\bibliographystyle{splncs04}
\bibliography{mybibliography}
%

\begin{comment}

\end{comment}

\end{document}